\documentclass{article}
\usepackage{spconf,amsmath,graphicx,hyperref}
\usepackage{cite}
\usepackage{amsmath,amssymb,amsfonts}
\usepackage{algorithmic}
\usepackage{graphicx}
\usepackage{multirow}
\usepackage{textcomp}
\usepackage{xcolor}
\usepackage{pgfplotstable}
\usepackage{tikz}
\usepackage{textcase}
\usepackage{float}
 \usepackage{booktabs}
\usepackage{pgfplots}
\usepackage{enumitem} 
\usetikzlibrary{pgfplots.groupplots}
\usetikzlibrary{decorations.pathreplacing}
\usetikzlibrary{pgfplots.fillbetween}
\pgfplotsset{compat=1.18}
\def\BibTeX{{\rm B\kern-.05em{\sc i\kern-.025em b}\kern-.08em
    T\kern-.1667em\lower.7ex\hbox{E}\kern-.125emX}}

\title{How I Built ASR for Endangered Languages with a Spoken Dictionary*\\
\thanks{\protect\lowercase{*\uppercase{a}dditional long-form speech-transcription is segmented and used to refine our models.}}
}
\name{Christopher Bartley, Anton Ragni}
\address{School of Computer Science, University of Sheffield, 211 Portobello, Sheffield S1 4DP, UK
\\
\{csjbartley1, a.ragni\}@sheffield.ac.uk}

\begin{document}
\ninept
\maketitle
\begin{abstract}
Nearly half of the world’s languages are endangered. Speech technologies such as Automatic Speech Recognition (ASR) are central to revival efforts, yet most languages remain unsupported because standard pipelines expect utterance-level supervised data. Speech data often exist for endangered  languages but rarely match these formats. Manx Gaelic ($\sim$2,200 speakers), for example, has had transcribed speech since 1948, yet remains unsupported by modern systems. In this paper, we explore how little data, and in what form, is needed to build ASR for critically endangered languages. We show that a short-form pronunciation resource is a viable alternative, and that 40 minutes of such data produces usable ASR for Manx ($<$50\% WER). We replicate our approach, applying it to Cornish ($\sim$600 speakers), another critically endangered language. Results show that the barrier to entry, in quantity and form, is far lower than previously thought, giving hope to endangered language communities that cannot afford to meet the requirements arbitrarily imposed upon them.
\end{abstract}
\begin{keywords}
low-resource, critically-endangered, automatic speech recognition
\end{keywords}
\section{Introduction}
\label{sec:intro}

More than 1,400 languages already have fewer than 1,000 speakers. At one loss every two weeks, these would disappear in roughly 54 years. Reversing this trend depends on community effort to strengthen transmission, expand learning opportunities, and raise visibility. Automatic Speech Recognition (ASR) has been shown to be central to these efforts \cite{DBLP:journals/corr/abs-2002-06675}, making oral archives more accessible to learners, educators, and researchers. Yet many endangered languages remain unsupported because they lack what is believed to be the necessary resource; an \textit{utterance-level} corpus of segmented speech–text pairs (5–15 s clips with per-segment transcripts). Nevertheless, developing speech technology for these languages remains one of the most pressing areas of research.

By contrast, endangered language communities tend to create transcribed speech in two different formats. The first we refer to as \textit{short-form}; isolated word or brief-phrase recordings, typically up to 5 seconds long. For example, Forvo \cite{forvo} is a crowdsourced pronunciation resource spanning 430+ languages, substantially more than modern multilingual speech datasets \cite{leeml, conneau2022fleursfewshotlearningevaluation, DBLP:journals/corr/abs-1912-06670}. The second is continuous recordings that run for minutes or even hours, such as radio broadcasts, interviews, and folklore (henceforth \textit{long-form}). In both cases, transcription text may be verbatim, but is often interspersed with translations, explanations, and other meta language. These “in-the-wild” data exist because they meet the most immediate needs of the community. One could argue that the challenge for speech technology is to adapt to such resources, rather than asking communities to build datasets whose immediate value to them is uncertain.

Languages with small speaker populations, such as Manx and Cornish, are often termed low-\textit{resource}, as though the challenge were simply a lack of data. Consequently, speech technologies for low-\textit{resource} languages tend to rely on crowdsourced utterance-level datasets \cite{DBLP:journals/corr/abs-1912-06670, khassanov-etal-2021-crowdsourced} and untold amounts of compute poured into massive pretrained models \cite{pratap2023scalingspeechtechnology1000}. However, these requirements are not always feasible for \textit{endangered} language communities, particularly for the smallest speaker populations whose languages are continually unrepresented in modern multilingual speech datasets and benchmarks, such as FLEURS \cite{conneau2022fleursfewshotlearningevaluation}, and CommonVoice \cite{DBLP:journals/corr/abs-1912-06670}.

Prior work in this area has explored extremely low-resource scenarios, such as the IARPA BABEL program, which used as little as three hours of manually aligned and segmented telephone speech \cite{gales2017low, thomas2016multilingual} to build ASR. Although unsupervised learning and zero- or few-shot adaptation have pushed requirements lower \cite{baevski2022unsupervisedspeechrecognition, liu2022endtoendunsupervisedspeechrecognition, zhang2023speakforeignlanguagesvoice}, success often depends on similarity to languages seen during pre-training. \cite{rouditchenko2023comparisonmultilingualselfsupervisedweaklysupervised}. In contrast, NLP research often alleviates supervision requirements in low-speaker settings by leveraging rudimentary resources common to many languages. For instance, bilingual dictionaries have proven highly effective in machine translation, both as the primary source of supervision \cite{duan-etal-2020-bilingual} and as components integrated directly into model architectures \cite{zhong2022lookupbilingualdictionaries}. Similarly, MTOB (Machine Translation from One Book) is a benchmark that asks large language models to translate Kalamang ($\sim$200 speakers) by reading a single grammar book \cite{tanzer2024benchmarklearningtranslatenew}. To the best of our knowledge, no one has explored using such rudimentary resources for ASR.

In this work, we ask \textit{What are the minimum requirements for developing ASR for critically endangered languages?} and \textit{Can short and long-form resources substitute for an utterance-level speech corpus?} We address these questions with a series of experiments and make the following contributions:  
\begin{enumerate}[label=\roman*)]
    \item We present the first ASR systems for two critically endangered languages, Manx and Cornish.
    \vspace{-0.1cm} 
    \item We show how 40 minutes of short-form speech can produce ASR ($<$50\% WER) for endangered languages.
    \vspace{-0.1cm} 
    \item With just 8 minutes of short-form speech, we train an initial ASR that can automatically segment long-form recordings for further ASR refinement. 
    \vspace{-0.1cm} 
    \item We achieve usable transcription technology ($<$25\% WER) with 40 additional minutes of segments.
    \vspace{-0.1cm} 
    \item We create the first utterance-level speech corpora for Manx (18 hours) and Cornish (39 hours).
\end{enumerate}

The rest of this paper is organised as follows. Section 2 discuss the background of this paper, section 3 presents a guide to data preparation, section 4 model devlopment and results, and section 5 concludes our findings.

\section{Background}
\label{sec:background}
\subsection{End-to-end (E2E) Models}
Fine-tuning end-to-end (E2E) models has emerged as the dominant paradigm for extending speech technologies to low-resource languages. Meta's MMS claims 'coverage' of 1,000+ languages \cite{pratap2023scalingspeechtechnology1000}. They define ‘coverage’ as the share of languages whose in-domain test sets achieve $<$5\% CER under a single model trained on New Testament readings across 1,107 languages (MMS-lab). Broad-domain ASR is shown only for a smaller subset of 102 languages. MMS is just one example of approaches that rely on extensive pre-training followed by fine-tuning. However, these methods come with broader downsides. Firstly, fine-tuning requires utterance-level speech data, which most endangered languages do not have. Secondly, popular fine-tuning toolkits like HuggingFace Transformers \cite{DBLP:journals/corr/abs-1910-03771}, limit control over model tuning to basic hyperparameters (e.g., learning rate, batch size), thus ignoring the benefits one could gain from finer-grained controls, such as incorporating specialised lexicons and external language models. Whilst it is possible to extend the standard recipes with these capabilities, doing so is difficult for non-experts. Lastly, fine-tuning E2E models is commonly believed to lead to state-of-the-art results universally, but this assumption does not hold for many low-resource languages \cite{klejch2025practitionersguidebuildingasr}.

\subsection{Hidden Markov Models (HMMs)}
In contrast, Hidden Markov Model (HMM) systems could be far more data and compute-efficient \cite{an2020catctccrfbasedasr}. For ASR tasks they are typically paired with a Gaussian mixture model (GMM) or a deep neural network (DNN). Their modularity requires finer-grained control over components such as acoustic modelling, pronunciation lexicons and language models, the latter of which enable greater integration of linguistic knowledge via web-based text data \cite{toshniwal2018comparisontechniqueslanguagemodel} (news, Wikipedia, digitised books), which are often more abundant than speech data for written languages. However, these systems are often regarded as difficult to use, with the dominant implementation in the Kaldi toolkit \cite{povey2011kaldi} demanding significant expertise and engineering effort. They too require utterance-level speech data, and while their modularity remedies some of the shortcomings of E2E modeling, their degree of success is often language-specific \cite{Gales2014SpeechRA}, depending heavily on the availability of linguistic resources like lexicons and text corpora.

\subsection{Community Efforts for Speech Technology}
Both paradigms inadvertently leave behind endangered languages; utterance-level speech–text corpora are rare to non-existent, and small speaker communities have limited capacity to produce them. However, hope resides in the fervour and sustained commitment that these communities have for their respective languages. Manx, for instance, has made a remarkable recovery from near extinction, growing its speaker base from just a few hundred in the 1960s to around 2,200 today. In the process, they have made public over 300 hours of educational material, podcasts, and recorded interviews. Similarly, Cornish revival efforts have rekindled an active speaker base of around 600, and prospective learners benefit from a variety of audible educational resources and media. Many remaining endangered languages likely have similar resources, and few--if any--have utterance-level recordings. A truly universal speech technology would make the most of these efforts and convert them into tools that serve the immediate needs of each community.

\section{Data Preparation}
\label{sec:dataprep}
\subsection{Short- and Long-form Speech}
To follow our approach, collect short-form speech and text files where the text corresponds exactly to the word or phrase spoken. Audio will typically be $\leq$ 2 seconds in length, and while longer clips are acceptable, the likelihood of a verbatim match tends to decline with duration. Any non-verbatim transcripts should be filtered out but kept aside for later steps. For sourcing these data, community “spoken dictionaries” and pronunciation sites are useful. Forvo \cite{forvo} has these for 430+ languages, and whilst valuable, this remains but a fraction of the world's $\sim$7,000 languages. They do, however, support crowdsourced additions to their collections. Failing this, we estimate it would take a few hours to collect a few hundred verbatim word or phrase clips with a prompt list and a microphone. For total duration and number of files, we show that as little as 8.5 minutes (433 files) can be effective. Our spoken dictionary collections for Manx and Cornish are summarised by the top block of Table \ref{tab:train_sets_duration}.

\begin{table}[t!]
\caption{Summary of Manx (gv) and Cornish (kw) data.}
\centering
\resizebox{1\columnwidth}{!}{%
\begin{tabular}{@{} l l l l l r r @{}}
  \hline
  Set & Lang. & Source & Domain & Sp. Style & \#Speakers & Dur. (mins) \\
  \hline
  \multicolumn{7}{l}{\hspace{1em}\textit{Short-form data}} \\
  A & gv    & Learnmanx    & Education & Careful & $>$10 & 102.13 \\
  B & gv    & Forvo        & Education & Careful & $>$10 & 15.95     \\
  C & kw    & Forvo        & Education & Careful & 9     & 8.5    \\
  \hline
  \multicolumn{7}{l}{\hspace{1em}\textit{Long-form data}} \\
  D & gv    & Learnmanx    & Education & Careful  & $>$10 & 183.68 \\
  E & gv    & Clilstore    & Religion  & Read     & 2     & 288.97 \\
  F & gv    & YouTube      & Interview & Conv.    & $>$10 & 261.81 \\
  G & gv    & Learnmanx    & Literary  & Read     & 1     & 237.30 \\
  H & kw    & Skeulantavas & Literary  & Read     & 1     & 2581.84 \\
  \hline
\end{tabular}%
}
\label{tab:train_sets_duration}
\end{table}

The second stage of our approach involves creating new speech segments (utterance-level) from long-form speech recordings ($\geq$ 30 seconds). Verbatim transcriptions of professional quality are preferred but not strictly required. So long as some portion of the text matches parts of the audio, new segments can be created. Gather as many recordings as possible that have accompanying text, whether in captions, descriptions, subtitles, or external documents. Choose sources that match your intended use. If the goal is to create usable transcription technology ($<$25\% WER) for read-speech (audiobook style) recordings, then prioritise  that data as domain-matched training usually gives the biggest gains. For a more general-purpose ASR, prioritise a mix of domains and speaking styles. In proceeding experiments, we investigate the amount of long-form data needed to adapt models to selected domains. The data we collect and use to build Manx and Cornish ASR come from publicly available web sources \footnote{Manx speech sources: \texttt{learnmanx.com}, \texttt{clilstore.eu}, \texttt{youtube.com/@learnmanx,@ManxNationalHeritage}} \footnote{Cornish speech sources: \texttt{forvo.com/languages/kw/}, \texttt{skeulantavas.com/audio}, \texttt{youtube.com/@Wikitongues}} and are summarised in the bottom block of Table \ref{tab:train_sets_duration}.


\subsection{Text Data}
\label{sec:text}
For written languages, web-based text data are often more plentiful than speech. For instance, \texttt{bible.com} has translations of the Bible in 2,300+ languages. Leveraging text data like these can substantially improve ASR by direct decoding or rescoring with external language models (LMs). Here we outline a simple approach to leveraging 4.8 million Manx words derived from web-based sources \footnote{Manx text sources: \texttt{manxradio.com}, \texttt{learnmanx.com}, \texttt{corpus.gaelg.im}, \texttt{gv.wikipedia.org}, \texttt{culturevannin.im}}. After collection, we derive a 72k list of Manx words by extracting and filtering a frequency-based wordlist from our text data and supplement rare words using a Manx dictionary \cite{dictionaryq_fockleyreen}. The resulting out-of-vocabularly (OOV) rate was 1.66\% across all Manx test sets. In the experiments that follow, we assess per domain how much gain a general LM trained on a mix of domains can provide when applied to ASR models (\ref{sec:manx}). For unwritten languages or languages where little to no text data is available, this stage is impractical. To simulate this case, for Cornish we use only the paired transcription text (no web text), and for Manx we report runs where the external LM is disabled.

\subsection{Test Sets}
In order to evaluate HMM and E2E modeling techniques, it was necessary to produce test sets, summarised by Table \ref{tab:subset_stats_duration}. For Manx, we selected recordings from three distinct domains and speaking styles: enunciated \textit{careful} speech from educational recordings intended for learners, traditional read-speech of religious texts, and spontaneous conversational speech from interview recordings. Recordings were randomly selected until their total length met or exceeded 15 minutes. We then cut each recording into timestamped audio intervals and matched each to its transcript line via Label Studio \cite{labelstudio}). A similar approach was taken for Cornish but for two test sets; a read-speech test from the same distribution as our training data and an out-of-domain spontaneous speech test (5 minutes) from a Youtube interview.

\begin{table}[t!]
\caption{Summary of test sets.}
\centering
\resizebox{1\columnwidth}{!}{%
\renewcommand{\arraystretch}{1.1}
\begin{tabular}{l l l l r r r}
  \hline
  \# & Lang. & Domain & Sp. Style & \#Speakers & \#Utts. & Dur. (mins) \\
  \hline
  T1 & gv & Education & Careful & $>$10 & 341 & 16.95 \\
  T2 & gv & Religion  & Read    & 2     & 278 & 16.67 \\
  T3 & gv & Interview & Conv.   & $>$10 & 279 & 16.03 \\
  T4 & kw & Podcast   & Read    & 1     & 131 & 14.60 \\
  T5 & kw & Interview & Conv.   & 1     & 37  & 5.18  \\
  \hline
\end{tabular}%
}
\label{tab:subset_stats_duration}
\end{table}

\subsection{Preprocessing}
\label{sec:preprocessing}
We converted all audio to a common format of 16-bit PCM, single-channel WAV, with a sampling rate of 16 kHz. We normalised all text, including transcription and unstructured text, by converting to uppercase and standardising whitespace. All punctuation was removed, except for intra-word hyphens and apostrophes (e.g. \textit{mother-in-law}) to preserve lexical meaning. We replaced diacritic markers with their canonical variants (e.g., \c{C}$\to$C; \'{E}$\to$E; \~{N}$\to$N). Numerals from 0--30 were expanded to their word forms using language-specific number lists from \texttt{omniglot.com}, which covers 2,246 languages. In some cases, a single transcription document covered multiple recordings. While not strictly necessary, we manually split these to enforce a one-to-one mapping between each long-form recording and its transcription to improve the chances of successful segmentation.

To use our long-form audio, we must first convert them into utterance-length segments. To do this, we use the Kaldi tooklit \cite{povey2011kaldi} and follow the forced alignment pipeline presented in LibriSpeech \cite{7178964} with two changes. Firstly, we adopt a Unicode graphemic lexicon \cite{7178960} as, like most other languages, Manx and Cornish do not have a phonemic pronunciation dictionary. Secondly, instead of using a pre-built acoustic model, we train a monophone\footnote{Training a robust triphone (context-dependent) system at this scale is challenging mainly due to data sparsity.} GMM–HMM on our short-form speech (Manx 102 mins, Cornish 8 mins) which is then used to bootstrap Viterbi forced alignment \cite{18626} of the long-form audio. We had an alignment success rate between 70-80\% for both systems. Substantially lower than this may point to missing/incomplete transcripts which require revision. Using our alignments, we train a triphone model. The Smith–Waterman alignment and segmentation steps remain the same as in LibriSpeech \cite{7178964}.

\begin{table}[t!]
\caption{\%WERs for Manx source comparison of training sets with different formats, speaker diversity, and quantity.}
\centering
\renewcommand{\arraystretch}{1.1}
\resizebox{1\columnwidth}{!}{%
\begin{tabular}{@{} l l l l c r r r @{}}
  \hline
  Quantity & Sets & Source & Format & \#Speakers & T1 & T2 & T3 \\
  \hline
  \multirow{4}{*}{16 mins} & A & Learnmanx & short-form & $>$10 & 53.88 & 79.26 & 85.96 \\
                           & B & Forvo & short-form & $>$10 & 97.44 & 96.19 & 99.30 \\
                           & D, E, F & multi & utterance-level & $>$10 & 53.15 & 62.18 & 79.79 \\
                           & G & Learnmanx & utterance-level & 1 & 75.92 & 85.39 & 97.56 \\
  \hline
  \multirow{3}{*}{102 mins} & A & Learnmanx & short-form & $>$10 & 42.73 & 73.80   & 81.11 \\
                            & D, E, F & multi & utterance-level & $>$10 & \textbf{34.94} & \textbf{37.89} & \textbf{59.31} \\
                            & G & Learnmanx & utterance-level & 1 & 79.37  & 87.48 & 97.79  \\
  \hline
\end{tabular}%
}
\label{tab:wer_splits}
\end{table}

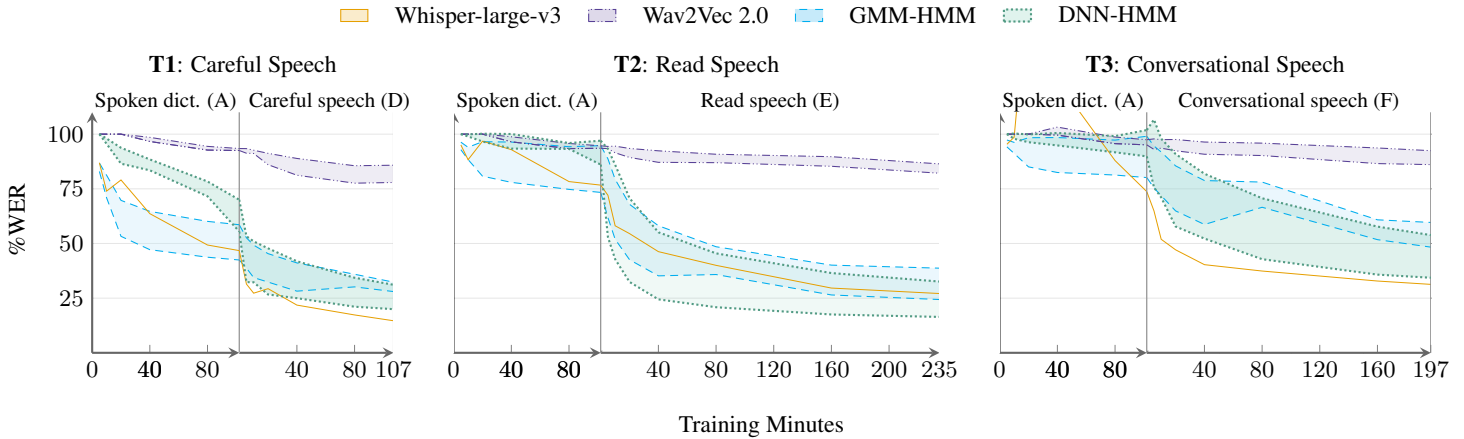
\begin{figure*}[ht] 
  \centering
  \begingroup
    \pgfplotsset{compat=1.18}

    \definecolor{whisperorange}{RGB}{230,159,0}
    \definecolor{wav2vecpurple}{RGB}{88,65,150}
    \definecolor{dnngreen}{RGB}{82,155,132}
    \definecolor{dnnfill}{RGB}{102,194,165}

    \pgfplotsset{every mark/.append style={/tikz/mark size=1pt}}

    \newlength\RightW   \setlength\RightW{4.6cm}        
    \newlength\UnitW    \setlength\UnitW{\dimexpr \RightW/240\relax}

    \newlength\Wleft    \setlength\Wleft  {\dimexpr 102\UnitW\relax}  
    \newlength\WcareR   \setlength\WcareR {\dimexpr 107\UnitW\relax}  
    \newlength\WreadR   \setlength\WreadR {\dimexpr 235\UnitW\relax}  
    \newlength\WconvR   \setlength\WconvR {\dimexpr 197\UnitW\relax}  

    \newlength\pairspacerwidth  \setlength\pairspacerwidth{0.8cm}     

    \pgfplotsset{
      intragap/.style={
        width=0, height=3.2cm,
        axis lines=none, ticks=none, clip=false,
        axis background/.style={fill=white}
      },
      pairgap/.style={
        width=\pairspacerwidth,  height=3.2cm,
        axis lines=none, ticks=none, clip=false,
        axis background/.style={fill=white}
      }
    }

    \noindent\makebox[\textwidth][c]{%
    \begin{tikzpicture}
      \begin{groupplot}[
        group style={group size=11 by 1, horizontal sep=0pt, vertical sep=0pt},
        height=3.2cm, 
        ymin=0, ymax=110,
        ytick={25,50,75,100},
        major tick length=0pt,
        minor tick length=0pt,
        ymajorgrids,
        every axis y grid/.style={draw=black!10, line width=.1pt},
        axis line style={color=black!60, line width=.8pt},
        tick label style={font=\small},
        xlabel={}, ylabel={},
        scale only axis=true 
      ]

\nextgroupplot[
  width=\Wleft,
  xmin=0, xmax=102,
  axis x line=bottom, axis y line=left,
  xtick={0,40},
  extra x ticks={40,80},
  extra x tick style={grid=none, tick align=inside, major tick length=3pt},
  legend to name=globallegend, legend columns=4,
  legend style={draw=none, column sep=8pt, font=\small}]
  \addplot[gray, forget plot] coordinates {(102,0) (102,110)};
  \addplot[cyan, densely dashed, name path=careful_LM1, forget plot]
    coordinates {(5,82.94)(10,70.90)(20,53.27)(40,47.17)(80,43.72)(102,42.45)};
  \addplot[cyan, densely dashed, mark options={solid,scale=0.8}, name path=careful_noLM1, forget plot]
    coordinates {(5,86.81)(10,81.22)(20,69.72)(40,64.68)(80,60.11)(102,58.58)};
  \addplot[fill=cyan!20, opacity=.35, draw=none, forget plot]
    fill between[of=careful_noLM1 and careful_LM1];
  \addplot[whisperorange, ultra thin, forget plot]
    coordinates {(5,86.92)(10,73.95)(20,79.05)(40,63.64)(80,49.30)(102,46.74)};
  \addplot[wav2vecpurple, densely dashdotdotted, forget plot]
    coordinates {(5,100)(10,100)(20,100)(40,96.65)(80,92.72)(102,92.53)};
  \addplot[wav2vecpurple, densely dashdotdotted, forget plot, name path global=caref_noLM1]
    coordinates {(5,100)(10,100)(20,100)(40,98.53)(80,94.41)(102,93.42)};
  \addplot[wav2vecpurple, densely dashdotdotted, forget plot, name path global=caref_LM1]
    coordinates {(5,100)(10,100)(20,100)(40,96.65)(80,92.72)(102,92.53)};
  \addplot[fill=wav2vecpurple, opacity=.10, draw=none, forget plot]
    fill between[of=caref_LM1 and caref_noLM1];
  \addplot[dnngreen, densely dotted, thick, forget plot, name path global=caref_noLM5]
    coordinates {(5,100)(10,97.99)(20,93.77)(40,88.44)(80,78.35)(102,70.01)};
  \addplot[dnngreen, densely dotted, thick, forget plot, name path global=caref_LM5]
    coordinates {(5,100)(10,95.59)(20,86.49)(40,83.46)(80,71.57)(102,55.48)};
  \addplot[fill=dnnfill, opacity=.20, forget plot, draw=none] fill between[of=caref_LM5 and caref_noLM5];

  \addlegendimage{legend image code/.code={\fill[whisperorange!20] (0,-0.08) rectangle (0.36,0.08);\draw[whisperorange, ultra thin] (0,-0.08) rectangle (0.36,0.08);}}
  \addlegendentry{Whisper-large-v3}
  \addlegendimage{legend image code/.code={\fill[wav2vecpurple!20] (0,-0.08) rectangle (0.36,0.08);\draw[wav2vecpurple, densely dashdotdotted] (0,-0.08) rectangle (0.36,0.08);}}
  \addlegendentry{Wav2Vec 2.0}
  \addlegendimage{legend image code/.code={\fill[cyan!20] (0,-0.08) rectangle (0.36,0.08);\draw[cyan, dashed] (0,-0.08) rectangle (0.36,0.08);}}
  \addlegendentry{GMM-HMM}
  \addlegendimage{legend image code/.code={\fill[dnnfill!20] (0,-0.08) rectangle (0.36,0.08);\draw[dnngreen, densely dotted, thick] (0,-0.08) rectangle (0.36,0.08);}}
  \addlegendentry{DNN-HMM}

\nextgroupplot[intragap]

\nextgroupplot[
  width=\WcareR,
  xmin=0, xmax=107,
  axis x line=bottom, axis y line=right, y axis line style={draw=none},
  xtick=\empty,
  extra x ticks={40,80, 107},
  extra x tick style={grid=none, tick align=inside, major tick length=3pt},
  yticklabels=\empty]
  \addplot[cyan, densely dashed,  name path=care_noLM] coordinates {(0,58.58)(5,52.60)(10,49.25)(20,45.35)(40,41.11)(80,35.93)(107,32.32)};
  \addplot[cyan, densely dashed, name path=care_LM]  coordinates {(0,42.45)(5,38.77)(10,34.40)(20,32.35)(40,28.20)(80,30.15)(107,28.01)};
  \addplot[fill=cyan!20, opacity=.35, draw=none] fill between[of=care_noLM and care_LM];
  \addplot[gray, forget plot] coordinates {(107,0) (107,110)};
  \addplot[gray, forget plot] coordinates {(0,0) (0,110)};
  \addplot[whisperorange, ultra thin, forget plot]
    coordinates {(0,46.74)(5,31.45)(10,27.27)(20,29.33)(40,21.82)(80,17.32)(107,14.66)};
  \addplot[wav2vecpurple, densely dashdotdotted, forget plot, name path global=caref_noLM2]
    coordinates {(0,93.42)(5,93.39)(10,92.65)(20,91.22)(40,88.89)(80,85.56)(107,85.79)};
  \addplot[wav2vecpurple, densely dashdotdotted, forget plot, name path global=caref_LM2]
    coordinates {(0,92.53)(5,91.28)(10,91.17)(20,86.04)(40,81.22)(80,77.58)(107,77.90)};
  \addplot[fill=wav2vecpurple, opacity=.10, draw=none] fill between[of=caref_LM2 and caref_noLM2];
  \addplot[dnngreen, densely dotted, thick, forget plot, name path global=caref_noLM4]
    coordinates {(0,70.01)(5,53.31)(10,50.97)(20,47.78)(40,41.87)(80,34.33)(107,31.11)};
  \addplot[dnngreen, densely dotted, thick, forget plot, name path global=caref_LM4]
    coordinates {(0,55.48)(5,32.32)(10,32.26)(20,26.67)(40,24.98)(80,21.08)(107,19.96)};
  \addplot[fill=dnnfill, opacity=.20, draw=none] fill between[of=caref_LM4 and caref_noLM4];

\nextgroupplot[pairgap]

\nextgroupplot[
  width=\Wleft,
  xmin=0, xmax=102,
  axis x line=bottom, axis y line=left,
  yticklabels=\empty,
  xtick={0,40,80},
  extra x ticks={40,80},
  extra x tick style={grid=none, tick align=inside, major tick length=3pt}]
  \addplot[gray, forget plot] coordinates {(102,0) (102,110)};
  \addplot[cyan, densely dashed, mark options={solid,scale=0.8}, name path=read_LM1]
    coordinates {(5,92.90)(10,88.38)(20,80.79)(40,77.95)(80,74.74)(102,73.36)};
  \addplot[cyan, densely dashed, mark options={solid,scale=0.8}, name path=read_noLM1]
    coordinates {(5,96.34)(10,93.98)(20,96.56)(40,96.38)(80,94.25)(102,94.77)};
  \addplot[fill=cyan!20, opacity=.35, draw=none] fill between[of=read_noLM1 and read_LM1];
  \addplot[whisperorange, ultra thin]
    coordinates {(5,95.07)(10,88.31)(20,96.77)(40,92.92)(80,78.31)(102,76.69)};
  \addplot[wav2vecpurple, densely dashdotdotted]
    coordinates {(5,100)(10,100)(20,100)(40,96.49)(80,93.39)(102,93.54)};
  \addplot[wav2vecpurple, densely dashdotdotted, forget plot, name path global=read_noLM3]
    coordinates {(5,100)(10,100)(20,100.07)(40,98.80)(80,95.74)(102,94.66)};
  \addplot[wav2vecpurple, densely dashdotdotted, forget plot, name path global=read_LM3]
    coordinates {(5,100)(10,100)(20,100)(40,96.49)(80,93.39)(102,93.54)};
  \addplot[fill=wav2vecpurple, opacity=.10, draw=none] fill between[of=read_LM3 and read_noLM3];
  \addplot[dnngreen, densely dotted, thick, forget plot, name path global=read_noLM5]
    coordinates {(5,100)(10,99.89)(20,100.19)(40,100.07)(80,95.93)(102,97.01)};
  \addplot[dnngreen, densely dotted, thick, forget plot, name path global=read_LM5]
    coordinates {(5,100)(10,98.95)(20,96.64)(40,93.39)(80,93.24)(102,86.08)};
  \addplot[fill=dnnfill, opacity=.20, draw=none] fill between[of=read_LM5 and read_noLM5];

\nextgroupplot[intragap]

\nextgroupplot[
  width=\WreadR,
  xmin=0, xmax=235,
  axis x line=bottom, axis y line=right, y axis line style={draw=none},
  yticklabels=\empty,
  xtick=\empty,
  extra x ticks={40,80,120,160,200, 235},
  extra x tick style={grid=none, tick align=inside, major tick length=3pt}]
  \addplot[gray] coordinates {(0,0) (1,1)};
  \addplot[cyan, densely dashed,  name path=read_noLM] coordinates {(0,94.77)(5,88.71)(10,79.33)(20,67.90)(40,58.11)(80,48.51)(160,40.10)(235,38.71)};
  \addplot[cyan, densely dashed, mark options={solid,scale=0.8}, name path=read_LM]  coordinates {(0,73.36)(5,61.81)(10,51.94)(20,42.56)(40,35.20)(80,35.80)(160,26.46)(235,24.36)};
  \addplot[fill=cyan!20, opacity=.35, draw=none] fill between[of=read_noLM and read_LM];
  \addplot[gray, forget plot] coordinates {(235,0) (235,110)};
  \addplot[gray, forget plot] coordinates {(0,0) (0,110)};
  \addplot[whisperorange, ultra thin]
    coordinates {(0,76.69)(5,72.04)(10,58.10)(20,54.51)(40,46.27)(80,40.00)(160,29.61)(235,27.11)};
  \addplot[wav2vecpurple, densely dashdotdotted, forget plot, name path global=read_noLM2]
    coordinates {(0,94.66)(5,94.39)(10,94.51)(20,93.46)(40,92.38)(80,90.81)(160,89.69)(235,86.40)};
  \addplot[wav2vecpurple, densely dashdotdotted, forget plot, name path global=read_LM2]
    coordinates {(0,93.54)(5,93.39)(10,91.48)(20,89.35)(40,87.11)(80,86.96)(160,85.31)(235,82.14)};
  \addplot[fill=wav2vecpurple, opacity=.10, draw=none] fill between[of=read_LM2 and read_noLM2];
  \addplot[dnngreen, densely dotted, thick, forget plot, name path global=read_noLM4]
    coordinates {(0,97.01)(5,93.27)(10,85.99)(20,70.70)(40,55.12)(80,45.48)(160,36.47)(235,32.59)};
  \addplot[dnngreen, densely dotted, thick, forget plot, name path global=read_LM4]
    coordinates {(0,86.08)(5,52.91)(10,42.83)(20,32.40)(40,24.44)(80,20.85)(160,17.53)(235,16.41)};
  \addplot[fill=dnnfill, opacity=.10, draw=none] fill between[of=read_LM4 and read_noLM4];

\nextgroupplot[pairgap]

\nextgroupplot[
  width=\Wleft,
  xmin=0, xmax=102,
  axis x line=bottom, axis y line=left,
  yticklabels=\empty,
  xtick={0,40,80},
  extra x ticks={40,80},
  extra x tick style={grid=none, tick align=inside, major tick length=3pt}]
  \addplot[gray, forget plot] coordinates {(102,0) (102,110)};
  \addplot[cyan, densely dashed, name path=conv_LM1]
    coordinates {(5,94.18)(10,91.19)(20,85.03)(40,82.47)(80,81.30)(102,80.14)};
  \addplot[cyan, densely dashed, mark options={solid,scale=0.8}, name path=conv_noLM1]
    coordinates {(5,97.32)(10,96.24)(20,98.37)(40,98.45)(80,97.28)(102,99.03)};
  \addplot[fill=cyan!20, opacity=.35, draw=none] fill between[of=conv_noLM1 and conv_LM1];
  \addplot[whisperorange, ultra thin]
    coordinates {(5,95.37)(10,98.61)(20,169.72)(40,124.15)(80,87.92)(102,73.73)};
  \addplot[wav2vecpurple, densely dashdotdotted]
    coordinates {(5,100)(10,100)(20,100)(40,99.57)(80,95.66)(102,95.11)};
  \addplot[wav2vecpurple, densely dashdotdotted, forget plot, name path global=conv_noLM2]
    coordinates {(5,100)(10,100)(20,100.04)(40,103.22)(80,98.68)(102,97.44)};
  \addplot[wav2vecpurple, densely dashdotdotted, forget plot, name path global=conv_LM2]
    coordinates {(5,100)(10,100)(20,100)(40,99.57)(80,95.66)(102,95.11)};
  \addplot[fill=wav2vecpurple, opacity=.10, draw=none] fill between[of=conv_LM2 and conv_noLM2];
  \addplot[dnngreen, densely dotted, thick, forget plot, name path global=conv_noLM5]
    coordinates {(5,100)(10,100.31)(20,99.88)(40,100.54)(80,99.11)(102,101.86)};
  \addplot[dnngreen, densely dotted, thick, forget plot, name path global=conv_LM5]
    coordinates {(5,100)(10,97.87)(20,96.31)(40,94.84)(80,91.70)(102,89.84)};
  \addplot[fill=dnnfill, opacity=.20, draw=none] fill between[of=conv_LM5 and conv_noLM5];

\nextgroupplot[intragap]

\nextgroupplot[
  width=\WconvR,
  xmin=0, xmax=197,
  axis x line=bottom, axis y line=right, y axis line style={draw=none},
  yticklabels=\empty,
  xticklabels=\empty,
  extra x ticks={40,80,120,160,197},
  extra x tick style={grid=none, tick align=inside, major tick length=3pt}]
  \addplot[gray] coordinates {(0,0) (1,1)};
  \addplot[cyan, densely dashed,  name path=conv_noLM] coordinates {(0,99.03)(5,94.57)(10,91.85)(20,85.49)(40,78.78)(80,78.08)(160,60.82)(197,59.62)};
  \addplot[cyan, densely dashed, name path=conv_LM]  coordinates {(0,80.14)(5,75.41)(10,71.88)(20,64.90)(40,58.69)(80,66.56)(160,51.75)(197,48.33)};
  \addplot[fill=cyan!20, opacity=.35, draw=none] fill between[of=conv_noLM and conv_LM];
  \addplot[whisperorange, ultra thin]
    coordinates {(0,73.73)(5,64.90)(10,51.92)(20,47.10)(40,40.31)(80,37.40)(160,32.83)(197,31.30)};
  \addplot[gray, forget plot] coordinates {(197,0) (197,110)};
  \addplot[gray, forget plot] coordinates {(0,0) (0,110)};
  \addplot[wav2vecpurple, densely dashdotdotted, forget plot, name path global=conv_noLM2]
    coordinates {(0,97.44)(5,97.83)(10,97.48)(20,97.56)(40,96.39)(80,95.89)(160,93.68)(197,92.44)};
  \addplot[wav2vecpurple, densely dashdotdotted, forget plot, name path global=conv_LM2]
    coordinates {(0,95.11)(5,93.02)(10,93.60)(20,92.40)(40,90.81)(80,90.22)(160,86.54)(197,86.07)};
  \addplot[fill=wav2vecpurple, opacity=.10, draw=none] fill between[of=conv_LM2 and conv_noLM2];
  \addplot[dnngreen, densely dotted, thick, forget plot, name path global=conv_noLM4]
    coordinates {(0,101.86)(5,106.67)(10,97.94)(20,90.92)(40,81.89)(80,70.67)(160,57.72)(197,53.84)};
  \addplot[dnngreen, densely dotted, thick, forget plot, name path global=conv_LM4]
    coordinates {(0,89.84)(5,75.68)(10,71.06)(20,57.76)(40,52.25)(80,42.82)(160,35.76)(197,34.33)};
  \addplot[fill=dnnfill, opacity=.20, draw=none] fill between[of=conv_LM4 and conv_noLM4];

\end{groupplot}

\node[anchor=south, yshift=-35pt]
  at ($(group c1r1.south west)!0.5!(group c11r1.south east)$) {Training Minutes};
\node[anchor=center, rotate=90, xshift=5pt]
  at ($(group c1r1.north west)!0.5!(group c1r1.south west)+(-1cm,0)$) {\%WER};
\node[anchor=south, font=\tiny, yshift=0pt]
  at ($(group c1r1.north west)!0.5!(group c11r1.north east)+(0,26pt)$) {\pgfplotslegendfromname{globallegend}};

\node (pair1title) [anchor=south, yshift=10pt]
  at ($(group c1r1.north west)!0.5!(group c3r1.north east)$)
  {\small \textbf{T1}: Careful Speech};
\node (pair2title) [anchor=south, yshift=10pt]
  at ($(group c5r1.north west)!0.5!(group c7r1.north east)$)
  {\small \textbf{T2}: Read Speech};
\node (pair3title) [anchor=south, yshift=10pt]
  at ($(group c9r1.north west)!0.5!(group c11r1.north east)$)
  {\small \textbf{T3}: Conversational Speech};

\node[anchor=south, yshift=-3pt, xshift=0pt] at (group c1r1.north) {\footnotesize Spoken dict. (A)};
\node[anchor=south, yshift=-3pt, xshift=5pt] at (group c3r1.north) {\footnotesize Careful speech (D)};

\node[anchor=south, yshift=-3pt] at (group c5r1.north) {\footnotesize Spoken dict. (A)};
\node[anchor=south, yshift=-3pt] at (group c7r1.north) {\footnotesize Read speech (E)};

\node[anchor=south, yshift=-3pt] at (group c9r1.north) {\footnotesize Spoken dict. (A)};
\node[anchor=south, yshift=-3pt] at (group c11r1.north) {\footnotesize Conversational speech (F)};

\end{tikzpicture}
}
\endgroup
\caption{\%WER performance of models trained on progressively larger amounts of Manx spoken dictionary data and new utterances from specific domains, assessed across careful, read, and conversational speech test sets. Shaded areas denote the gain from external LM integration (top vs.\ bottom line).}
\label{fig:splits-global-labels-final}
\end{figure*}

\section{Experiments}
\label{sec:experimentalsetup}
We begin by establishing baselines for short-form and utterance-level Manx data to investigate source diversity, form, and quantity. Building on this, we ask how little supervision is needed to achieve usable transcription technology ($<$25\% WER), with and without external LMs. Finally, we extend our approach to Cornish to assess generality to other endangered languages when text is scarce.

\subsection{Source Comparison}
\label{sec:sourcecomparison}
\textbf{Setup} -- Using the LibriSpeech recipes from the Kaldi toolkit \cite{povey2011kaldi}, we train a single GMM–HMM baseline with speaker-adaptive transforms\footnote{LDA+MLLT, then SAT/fMLLR. Speaker adaptive features are estimated per utterance in our case because speaker identities are unavailable}. During decoding, we use a word-based four-gram language model trained on 4.8M Manx words, using the wordlist described in \ref{sec:text}. These remain constant and evaluation is carried out on the same three test sets (T1–T3). We then vary data format (short- vs. utterance-level), speaker diversity ($>$10 vs. 1), and quantity (16 vs. 102 mins). At 16 minutes, we compare random samples of single- and multi-speaker utterance-level data with short-form sets from \textit{LearnManx} and \textit{Forvo}, the latter included because its clips are shorter on average and its speaker and lexical diversity are lower (despite $>$10 speakers, 1 speaker accounts for most of this data). At 102 minutes, we further compare single- and multi-speaker utterance-level data with the whole \textit{LearnManx} short-form set.

\par\addvspace{.5\baselineskip}

\noindent\textbf{Results} -- Our 102-minute, multi-speaker utterance-level setting achieved the best results across all three test set, followed closely for T1 by the \textit{LearnManx} spoken dictionary. We suspect this may be due to some speaker overlap between these sets. Source diversity in speakers, styles and domains is the the most determining factor; the single-speaker utterance-level systems were outperformed at both quantities by the \textit{LearnManx} spoken dictionary systems, and the much less diverse Forvo system performed markedly worse than all others at 16-minutes. Quantity had the least impact, with the single-speaker system performing worse with more data. 

\subsection{Minimum Supervision Requirements}
\label{sec:manx}
\textbf{Setup} -- Using 5, 10, 20, 40, 80, 102-minute subsets of the short-form \textit{LearnManx} set, we train and assess models across our careful, read, and conversational speech test sets. Once the short-form data are exhausted, we branch each model into three domain-specific runs. For each branch, we add multi-speaker utterance-level data from a single domain (careful, read, or conversational) in progressively larger subsets, assessing each against its in-domain test set. We evaluate two HMM systems—a speaker adaptive GMM–HMM and a TDNN-F model trained with LF-MMI from scratch (no GMM bootstrap)—and two E2E models (Wav2Vec 2.0 and Whisper-large-v3). We fine-tune our pre-trained models according to the SpeechBrain \cite{speechbrain} LibriSpeech recipes (LoRA). To investigate how well each model leverages text data, we incorporate the LM from the previous experiment. For the HMMs and wav2vec 2.0, the LM is integrated directly during decoding, whereas for Whisper it is used for n-best list rescoring.

\par\addvspace{.5\baselineskip}

\noindent\textbf{Results} -- From the spoken-dictionary data alone (Figure \ref{fig:splits-global-labels-final}), only the GMM–HMM (with LM) and Whisper generalise meaningfully to careful speech ($<$50\% WER). The GMM-HMM was the most impressive here, achieving $<$50\% WER on careful speech after just 40 minutes of short-form data. Adding domain-matched utterances drives clear specialisation. Whisper finishes best on careful (14.66\% WER) and conversational (31.30\% WER) speech, but is overtaken on read speech by the DNN-HMM with external LM integration (16.41\% WER). The shaded areas for each show that our LM contributes most to the HMMs, typically $\sim$20 absolute WER points (largest on read/conversational), whereas it resulted in 2 point average gain for wav2vec 2.0 in direct decoding mode and none for Whisper in n-best list rescoring mode.

\subsection{Cornish}
\textbf{Setup} -- Guided by these findings, we apply a similar approach to Cornish using 8.5 minutes of short-form data from Forvo. Despite the issues with this source highlighted in \ref{sec:sourcecomparison}, we are able to use it to perform the same alignment process described in \ref{sec:preprocessing}. As a result, we create a 39 hour speech corpus from the long-form Cornish resources in Table \ref{tab:train_sets_duration}. These involve only one speaker, which we expect to be detrimental to generalisation, however no other long-form resources are available at this time. To simulate a scenario where text resources are scarce, we neglect the use of an external LM and instead fine-tune whisper-large-v3 on this corpus. Evaluation is performed on a 14-minute in-domain read speech set (T4) and a short out-of-domain interview-style recording (T5).

\par\addvspace{.5\baselineskip}

\noindent\textbf{Results} -- Our Whisper system achieves an in-domain test set performance of 7.72\% WER (1.65\% CER) which passes Meta's MMS threshold of "coverage" \cite{pratap2023scalingspeechtechnology1000}. However, it also achieves 72.55\% WER (34.17\% CER) on the out-of-domain test, indicating severe domain and speaker mismatch. We suspect that this discrepancy comes mainly from the fact that the 39 hour corpus used for training all comes from a single speaker. However, research from the IARPA BABEL program shows that ASR systems with out-of-domain WERs of $\sim$70\% are still useful for tasks such as Keyword Spotting \cite{7078630}.

\section{Conclusion}
\label{sec:conclusion}
ASR support for endangered languages remains sparse, principally due to a lack of utterance-level supervision. Addressing this technological gap, we have shown that usable ASR can be attained for endangered languages that have no utterance-level supervised data by leveraging rudimentary resources such as a spoken dictionary. We show that just 40 minutes of spoken dictionary data from multiple speakers can produce a viable ASR baseline ($<$50\% WER). Our results highlight the flexibility of HMMs in integrating short- and long-form supervision and text information to deliver usable ASR (both HMM and E2E based) from unconventional resources. On the other hand, Whisper generalises markedly better to conversational speech, although the same may not be true for other E2E models such as wav2vec 2.0. Ultimately, this paper has shown that developing speech technology for low-resource and endangered languages does not require utterance-level corpora as a starting point. Collecting new and clean resources is an important task, but future work would do well to make sense of the data that already exist and continue to be created. Doing so may mean difference to the most urgent cases of language endangerment.

\vfill\pagebreak

\bibliographystyle{IEEEbib}
\bibliography{refs}

\end{document}